\pdfoutput=1

\documentclass[11pt]{article}
\usepackage{colortbl}
\usepackage[final]{acl}
\usepackage{xltabular}
\usepackage{longtable}
\usepackage{booktabs} 
\usepackage{array}    

\usepackage{xcolor}

\usepackage{amsmath}
\usepackage{pdflscape}
\usepackage{pdfpages}

\usepackage{circledsteps}
\usepackage{xspace}
\usepackage{times}
\usepackage{latexsym}

\usepackage[T1]{fontenc}

\usepackage[utf8]{inputenc}

\usepackage{microtype}

\usepackage{inconsolata}

\usepackage{graphicx}

\title{Natural Language Processing in Support of Evidence-based Medicine: A Scoping Review}

\author{
 \textbf{Zihan Xu\textsuperscript{1,*}},
 \textbf{Haotian Ma\textsuperscript{1,*}},
 \textbf{Gongbo Zhang\textsuperscript{2}},
 \textbf{Yihao Ding\textsuperscript{3}},
\\
 \textbf{Chunhua Weng\textsuperscript{2}},
 \textbf{Yifan Peng\textsuperscript{1}}
\\
\\
 \textsuperscript{1}Weill Cornell Medicine,
 \textsuperscript{2}Columbia University,
 \textsuperscript{3}University of Sydney
\\
 \small{
   \textbf{Correspondence:} \url{yip4002@med.cornell.edu}
 }\\
 \textsubscript{*}\small{Authors contributed equally}
}

\begin{document}
\maketitle
\begin{abstract}
Evidence-based medicine (EBM) is at the forefront of modern healthcare, emphasizing the use of the best available scientific evidence to guide clinical decisions. Due to the sheer volume and rapid growth of medical literature and the high cost of curation, there is a critical need to investigate Natural Language Processing (NLP) methods to identify, appraise, synthesize, summarize, and disseminate evidence in EBM.
This survey presents an in-depth review of 129 research studies on leveraging NLP for EBM, illustrating its pivotal role in enhancing clinical decision-making processes. The paper systematically explores how NLP supports the five fundamental steps of EBM -- Ask, Acquire, Appraise, Apply, and Assess. The review not only identifies current limitations within the field but also proposes directions for future research, emphasizing the potential for NLP to revolutionize EBM by refining evidence extraction, evidence synthesis, appraisal, summarization, enhancing data comprehensibility, and facilitating a more efficient clinical workflow.
\end{abstract}

\section{Introduction}
Evidence-based medicine (EBM) is at the forefront of modern healthcare, emphasizing the use of the best available scientific evidence to guide clinical decisions~\cite{sackett1996evidence}. By integrating clinical expertise, patient values, and the most up-to-date research data, EBM facilitates healthcare decisions by patients and the general public, clinicians, guideline developers, administrators, and policymakers \cite{Mehta2022-at, Kwaan2012-nz, Van-de-Vliet2023-fr}. 

The foundation of EBM heavily relies on comprehensive research data from detailed textual sources such as clinical trial publications, cohort studies, and case reports \cite{Blunt2022PyramidSchema, 5A}. Navigating this evidence hierarchy necessitates the use of advanced Natural Language Processing (NLP) techniques, which are crucial for streamlining literature searches and extracting PICO (Patient/Population, Intervention, Comparison, Outcomes) elements \cite{Peng2023-zx, Nye2018-zs}. From the early utilization of statistical machine learning \cite{Arora2019} and recurrent neural networks \cite{Guan2019}, there has been a significant shift towards more advanced technologies such as transformer-based frameworks and large language models (LLMs). These modern approaches employ self-supervised pretraining and instruct-tuning \cite{Rohanian2024} to capture domain-specific knowledge \cite{Kalyan2022}, enhancing the accuracy and scalability of medical information processing \cite{Thirunavukarasu2023}. Particularly, the recent advancements in LLMs have further propelled NLP capabilities within EBM, excelling in more complex tasks such as appraising and synthesizing evidence \cite{Gorska2024-qy}, differentiating and ranking evidence \cite{380}, generating human-like responses, answering complex clinical questions \cite{Shiraishi2024-la}, and identifying relevant clinical trials \cite{Devi2024-ha}.

Despite these significant advancements, a comprehensive review summarizing NLP development and applications in EBM is still in demand. 
This paper seeks to fill the gap by offering a thorough review of essential NLP tasks in EBM, with a focus on evidence generation, such as evidence retrieval, extraction, synthesis, and summarization, as well as evidence adoption and evidence-based research, such as question-answering, clinical trial design and identification, and other cutting-edge studies across various clinical specialties.  

Furthermore, we outline key benchmarks to facilitate the development of future NLP models. Finally, we explore several potential avenues for future research. To better support both clinicians and researchers in making more informed clinical decisions and producing more comprehensive review literature, we have made these resources publicly available.\footnote{\url{https://github.com/bionlplab/awesome-nlp-in-ebm}}


\section{Scope and Literature Selection}
\begin{figure}
    \centering
    \includegraphics[width=\linewidth]{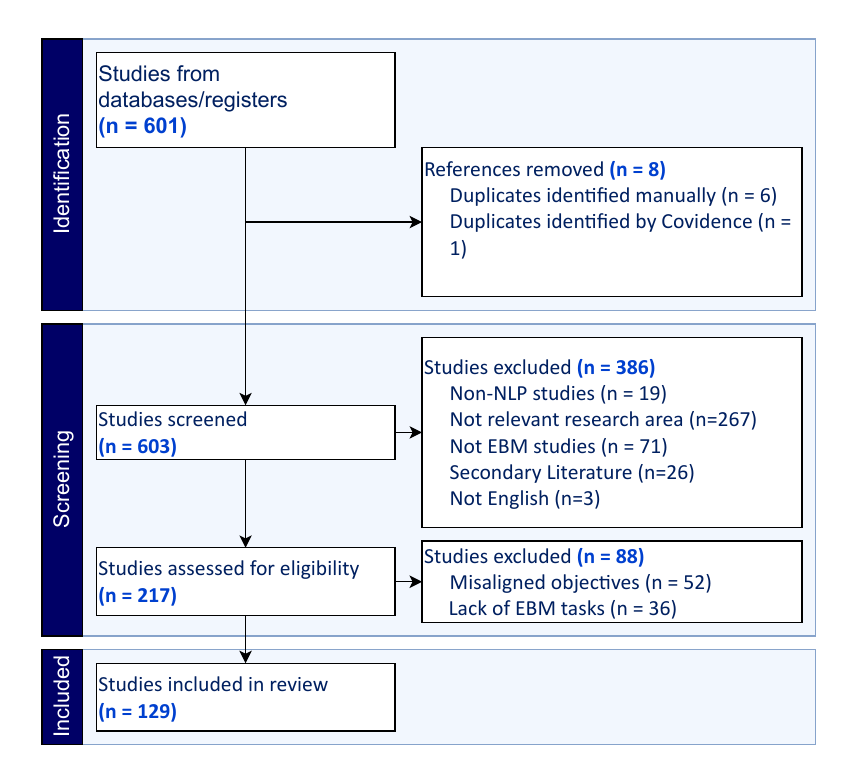}
    \caption{PRISMA flow diagram.}
    \label{fig:overall}
\end{figure}

Our scoping review adheres to the Preferred Reporting Items for Systematic Reviews and Meta-Analyses (PRISMA\footnote{\url{https://www.prisma-statement.org/}}) guidelines, as illustrated in Figure~\ref{fig:overall}. 

\subsection{Information sources}\label{information-sources}

We searched 4 databases, including PubMed\footnote{\url{https://pubmed.ncbi.nlm.nih.gov/}}, IEEE Xplore\footnote{\url{https://ieeexplore.ieee.org/}}, ACM Digital Library\footnote{\url{https://dl.acm.org/}}, and ACL Anthology\footnote{\url{https://aclanthology.org/}}. The search included studies from the past 5 years, spanning 2019 to 2024.

\subsection{Search strategy}\label{search-strategy}

Our search strategy was meticulously designed to capture the most relevant studies at the intersection of NLP and EBM (Supplementary File \ref{sec:query}). We targeted key NLP concepts and technologies by including terms such as `\textit{natural language processing}', `\textit{language model}', `\textit{large language model}', `\textit{computational linguistics}', `\textit{information extraction}', `\textit{information retrieval}', `\textit{clinical trial retrieval}', `\textit{text summarization}', `\textit{question answering}', `\textit{sentence segmentation}', `\textit{named entity recognition}', `\textit{tokenization}' and the abbreviations like `NLP' and `LLM'. In the domain of EBM, we included terms like `\textit{Evidence-Based Medicine}', `\textit{Evidence-Based Practice}', `\textit{Clinical Trial}' and their abbreviations like `EBM' and `EBP', also limited to appearances in the title or abstract. We used the Boolean operator to combine any word from the NLP domain and any work from the EBM domain in our search terms.

\subsection{Study selection and metadata extraction}\label{study-selection-and-metadata-extraction}

The references of all eligible studies were imported into Covidence\footnote{\url{https://www.covidence.org}}, and duplicates were removed. We then screened the articles by title and abstract. 
Inclusion criteria were defined as (1) Studies published in English, (2) research applying NLP techniques specifically for EBM, and (3) Studies focusing on applications for humans. 
Exclusion criteria were defined as (1) articles unrelated to NLP for EBM, (2) non-English publications, and (3) secondary literature such as systematic reviews, retracted papers, survey papers, case studies, and descriptive papers lacking experimental results. 

After the screening, the metadata was extracted from each paper, including models, disease, tasks involved, results, and limitations. 
Two annotators cross-verified the study selection and metadata extraction processes and consulted a third in cases of disagreement. 

\subsection{Study Statistics}
From an initial pool of 601 papers retrieved from databases and 9 additional sources, we removed 8 duplicates. Subsequently, 386 papers were excluded during the initial screening based on predefined exclusion criteria, and 88 more were removed during full-text screening due to misaligned objectives or lack of relevance to EBM tasks. Ultimately, 129 studies met the inclusion criteria and form the basis of this review, with detailed metadata provided in Supplementary Table~\ref{tab:studies}.

\begin{figure}
    \centering
    \includegraphics[width=\linewidth]{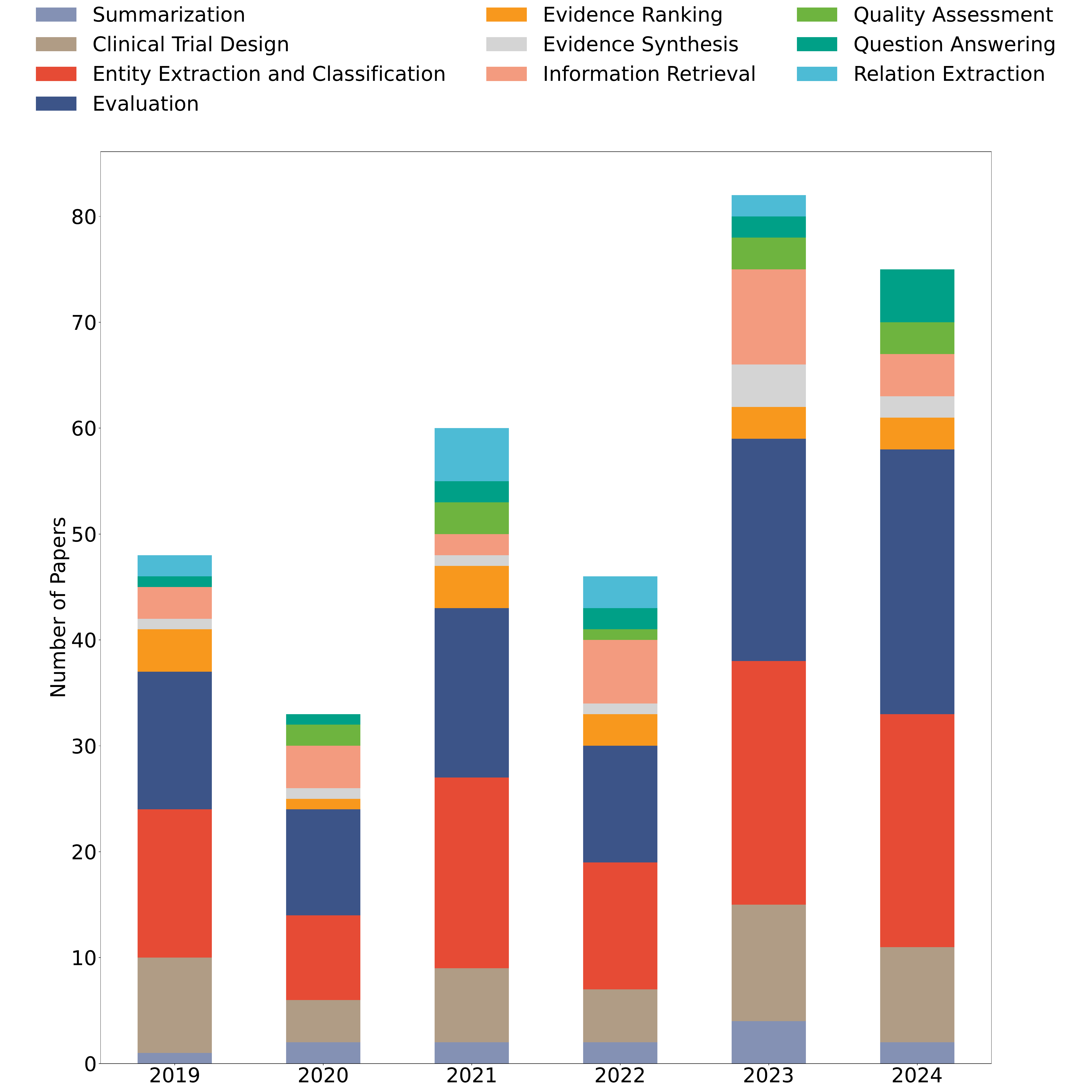}
    \caption{Distribution of papers in different EBM tasks over time. The color schema is the same as Supplementary Table ~\ref{tab:studies}.}
    \label{fig:distribution}
\end{figure}

Figure~\ref{fig:distribution} illustrates the distribution of research papers across different years (2019–2024) and their corresponding NLP tasks. There has been a rapid growth of papers over the years, peaking in 2023. The most common tasks throughout the years are Entity Extraction, Classification, and Evaluation, showing their foundational role in NLP for EBM research. Emerging tasks like Question Answering and Quality Assessment have appeared more prominently in recent years, reflecting evolving research directions. 

\section{NLP Techniques for EBM}\label{results}
The entire EBM process consists of five steps, commonly referred to as the `5A's: \textbf{Ask}, \textbf{Acquire}, \textbf{Appraise}, \textbf{Apply}, and \textbf{Assess}~\cite{5A}. NLP can be leveraged at each step to enhance the process (Table \ref{tab:tasks}). For example, in the \textbf{Ask} step, clinicians or patients formulate precise clinical questions to address specific healthcare concerns. During the \textbf{Acquire} step, NLP can be employed to extract evidence, often leveraging the PICO framework. In the \textbf{Appraise} step, NLP tools can assist in evaluating and ranking the quality, validity, and relevance of the retrieved information to ensure its applicability to clinical decision-making. For the \textbf{Apply and Assess} steps, NLP can streamline the design and identification of relevant clinical trials and facilitate their integration into practice, enabling continuous assessment and refinement of patient care strategies. Detailed trends and advancements for each step of the EBM process are discussed in the following sections.


\begin{table}
\rowcolors{2}{}{black!10!white}
\small
\begin{tabularx}{\columnwidth}{l
>{\raggedright\arraybackslash}X
>{\raggedright\arraybackslash}X}
\toprule
EBM cycle & Description & NLP tasks\\
\midrule
Ask & Search \& select studies & Question answering, Information retrieval\\
Acquire & Collect data & Named entity recognition and normalization, Relation extraction\\
Appraise & Examine relevance, validity, and results & Quality assessment, Evidence ranking and screening, Evidence synthesis, Evidence summarization\\
Apply \& Asses &  Apply EBM in practice and research and evaluate their effectiveness & Clinical trial identification and design, Question Answering, Domain-specific applications\\
\bottomrule
\end{tabularx}
\caption{Mapping of EBM cycle to corresponding NLP tasks.}
\label{tab:tasks}
\end{table}


\section{Ask - Searching \& Selecting Studies}\label{evidence-extraction-and-retrieval}

EBM can help researchers and clinicians draft a successful systematic review. After the scope and questions have been determined, the first step is to search for studies to include in the reviews and ensure they remain up to date. 

This step is typically achieved using NLP-based information retrieval techniques, which extract relevant information from large text corpora based on user queries. Early heuristic methods involved structured, keyword-based queries to retrieve articles from repositories like MEDLINE or PubMed. These methods, while foundational, are limited by the high cost of expert annotation, maintenance, and domain sensitivity \cite{neveol2011semi}. Despite these limitations, recent methods often rely on predefined rule-based strategies, e.g., SR[pt] and CQrs \cite{55}, to filter and compare the retrieved results for systematic reviews. In addition, while statistical machine learning and context-aware models \cite{Kamath2021-ux,182} have been widely adopted, they often lack scalability and struggle with less representative text embeddings. 

Recent advancements are leaning towards transformer-based deep learning frameworks \cite{Ramprasad2023AutomaticallySummarizing,5} due to their scalability and the ability to integrate medical ontologies, improving domain-specific text representation through self-supervised pretraining. For example, \cite{Lokker2023-tx} used BioBERT's \cite{lee2020biobert}  embeddings and attention mechanisms to improve query representation and biomedical literature retrieval in clinical practice. Furthermore, the integration of generative AI models has advanced literature retrieval despite challenges like hallucination. For example, \citet{69} compared Microsoft Bing AI and ChatGPT in accelerating the systematic literature search for a clinical review on Peyronie disease treatment, finding both can speed up the search process.

\section{Acquire - Collecting Data}

EBM is designed to identify all studies relevant to their research questions and synthesize data regarding the study design, risk of bias, and results. Therefore, the findings of EBM heavily depend on decisions about which data from these studies are presented and analyzed. The data collected should be accurate, complete, and accessible for future review, updates, and data-sharing purposes. Here we describe NLP approaches used to extract data directly from journal articles and other studies' reports.

\subsection{Entity Extraction and Normalization}\label{entity-extraction-and-classification}

Initially, entity (e.g., PICO) extraction relied on rule-based approaches, which utilize predefined lexical, syntactic, and contextual rules for extracting entities from clinical trial data \cite{316,292}. These methods are simple, transparent, and customizable, making them practical for high-precision tasks in structured contexts. Although they face challenges with complex or ambiguous data, their interpretability and ease of adaptation remain valuable for PICO extraction \cite{54}. 

RNN/LSTM-based frameworks lacked long-term memory capabilities. Nevertheless, they have been used for sequential sentence classification to enhance context utilization and improve classification accuracy in unstructured or less structured medical abstracts~\cite{Jin2018-tj}. 

The current trend is towards the dominance of transformer-based frameworks due to their domain-aware pertaining benefits. For instance, models such as SciBERT and PubMedBERT have been specifically developed for extracting `Intervention' (`I' in PICO) \cite{Tsubota2022-qf}, SrBERT~\cite{Aum2021-fp} for classifying articles into ``included' or ``excluded'' categories based on predefined inclusion criteria. 

\subsection{Relation Extraction}\label{relation-extraction} 

Following the identification of PICO elements, relation extraction approaches can be used to link these elements within studies. 

Initially, rule-based and machine-learning methods were used to extract meaningful relationships from medical literature~\cite{Alodadi2019, 292}. By 2021, transformative methodologies were developed, integrating deep learning frameworks like BERT and Augment Mining (AM). For example, srBERT built-in \cite{Aum2021-fp}, identified key elements and defines interrelations from the titles of articles. \citet{Stylianou2021-ek} classified the relationships between argumentative components within the texts, such as claims and evidence. Their relationships were labeled as `supporting' or `opposing'. 

In the systematic review process, understanding the connections between different study results can influence the review outcomes. However, besides systematic reviews, automated relation extraction has shifted towards more structured approaches, such as schema-based relation extraction. For example, \citet{Sanchez-Graillet2022-up} utilized a richly annotated corpus that aligns with the C-TrO ontology. Complementing these advances, graph-based approaches offer a novel way to encode complex relationships between clinical entities. A knowledge graph is a structured representation of information where entities (e.g., symptoms, treatments, drugs) are represented as nodes and their relationships as edges. Graph-based approaches have emerged as an effective method to encode relationships. For example, a knowledge graph was used to organize and visualize relationships among clinical trial entities such as symptoms, treatments, and drug outcomes by structuring data into nodes and edges \cite{Pan2021-eg}.

\section{Appraise, Synthesize, and Summarize Evidence}\label{evidence-synthesis-and-appraisal}

This task screens the included studies for risk of bias and appraises them for quality to ensure that healthcare decisions are informed by the most reliable and relevant evidence. Once the appraisal is complete, the next step is synthesizing evidence by combining findings from multiple studies, often using meta-analyses. Finally, these synthesized insights are summarized into concise, actionable conclusions.

\subsection{Quality Assessment}\label{quality-assessment}
Developing tools to assess evidence is crucial in EBM, such as the fully automated tool that combines machine learning and rule-based techniques by \citet{Brassey2021-ua}. It assessed the evidence from randomized clinical trials and systematic reviews by sentiment analysis, indication of bias, and sample size calculation, and used them to estimate the potential effectiveness of the intervention. Besides, deep learning models such as BERT \cite{Devlin2018-wo} have been used to evaluate the quality of evidence by analyzing article titles and abstracts. For example, different variations like BioBERT, BlueBERT, and BERT\textsubscript{BASE} were fine-tuned to classify the articles based on their adherence to methodological quality criteria \cite{Lokker2023-tx}. 

\subsection{Evidence Ranking and Screening}\label{evidence-ranking-and-screening}

After the quality assessment, the next step is to screen and rank the evidence. Several ranking methods are available, with statistical-based methods being among the earliest used. For example, \citet{Norman2019-nm} developed a method to rank references by their likelihood of relevance. Compared with randomized screening, their study showed that prioritization methods (with technological assistance) allow for fewer studies to be screened while still producing reliable results, which effectively reduces both the time and cost associated with the screening process.
\citet{Rybinski2020-rs} introduced the platform A2A, which used Okapi Best Match 25 (BM25) that assigned scores to documents based on term frequency and document length and Divergence from Randomness (DFR) that quantified informativeness as the divergence of a term's distribution from randomness for document ranking. Additionally, machine learning methods are implemented. \citet{Rybinski2020-mz} designed a search system with a simple query formulation strategy for initial ranking and used pre-trained BERT models (SciBERT, BioBERT, and BlueBERT) for re-ranking in clinical trial searches, which improved the robustness. 

\subsection{Evidence Synthesis}\label{findings-synthesis}

Evidence synthesis combines data from included studies to draw conclusions about a body of evidence. While the most common method used is meta-analysis, which statistically combines results from studies to estimate overall effect sizes, NLP-based approaches have also been applied to synthesize studies or findings. \citet{Mutinda2022-jx} proposed a method to reproduce meta-analysis, computing summary statistics (e.g., risk ratio) and visualizing results using forest plots by extracting and normalizing PICO elements from breast cancer randomized controlled trials. However, this method is built on a small amount of data. \citet{Gorska2024-qy} developed a system to continuously update summary statistics from key publications, further improving the meta-analysis process. However, only binary outcomes were supported in both methods, limiting the applicability to broader meta-analysis needs. 
Besides meta-analysis, EvidenceMap \cite{Kang2023-ei} effectively synthesized medical findings by employing a structured and hierarchical representation comprising Entities, Propositions, and Maps that enhances the interpretability and retrievability of evidence through its sophisticated semantic relational retranslation.

\subsection{Evidence Summarization}\label{evidence-summarization}

Finally, EBM must present a clear statement of findings or conclusions to help people make better-informed decisions and increase usability. This summary should include information on all important outcomes, evidence certainty, and the intervention's desirable and undesirable consequences.


From an NLP technical perspective, evidence summarization uses extractive and abstractive strategies. Extractive summarization selects the most important sentences from the original text. \citet{Gulden2019-dv} generated a new dataset from \url{clinicaltrials.gov} to test various algorithms (e.g., LexRank, TextRank, and Latent Semantic Analysis), identifying TextRank as the best performer in creating summaries directly from the source texts without altering the original wording. However, these algorithms suffered from inefficiency and high computational complexity when processing large datasets. \citet{Sarker2020-uh} developed a lightweight system that leverages Maximal Marginal Relevance (MMR) and pre-trained word embeddings trained on PubMed and PMC texts to integrate semantic relevance and reduce redundancy. Similarly, \citet{Xie2022-oa} proposed a knowledge infusion training framework called KeBioSum, which incorporated PICO into pre-trained language models (PLMs). It utilized lightweight knowledge adapters to reduce computational costs while improving semantic understanding and contextual representation.

Abstractive summarization focuses on the most critical information and creates new text for the summary; usually, more advanced techniques are used. \citet{Lalitha2023-tk} have implemented sophisticated techniques such as neural network-based model T5 (Text-to-Text Transfer Transformer), BART (Bidirectional Auto-Regressive Transformer), and PEGASUS (Pre-training with Extracted Gap-sentences for Abstractive Summarization Sequence-to-sequence) to resolve the challenge of obtaining useful information from a vast amount of clinical documents. 

With the increased demand for user-interacted summarization, \citet{Ramprasad2023-br} presented TrialsSummarizer, a system that helps automate summarizing the most relevant evidence in a set of randomized controlled trials by a multi-headed architecture, enabling each token in the generated summary to be explicitly linked to specific input aspects (e.g., population, intervention, or outcome). It introduces template-infilling capabilities, allowing users to correct or adjust generated summaries dynamically. Moreover, the application of LLMs has evolved to address these tasks with growing precision and depth. \citet{Hamed2023-jr} explored ChatGPT's capabilities in synthesizing diabetic ketoacidosis (KDA) guidelines by comparing, integrating, and abstracting content. \citet{Unlu2024-oj} employed a Retrieval-Augmented Generation (RAG) framework with GPT-4 for generating responses to clinical trial eligibility questions based on retrieved patient data. Furthermore, TriSum \cite{Jiang2024-zi} stood out by using structured rationale-based abstractive summarization, where large language models generate aspect-triple rationales that are distilled into smaller models through a dual-scoring selection mechanism and curriculum learning. 

\section{Apply and Assess: adoption, refinement, and research}\label{application-specialty}

Transitioning from the evidence generation and synthesis, the next critical step is its adoption and refinement, facilitated by an `Evidence-based Research' approach. Adoption and refinement are crucial to consistently reassessing and enhancing clinical evidence, particularly when existing evidence gaps lead to unmet needs of clinicians and patients. Evidence-based research further ensures that these gaps inform future clinical studies. 
%
Here, we summarize several applications identified from our literature review that align with this topic.

\subsection{Specialty-specific adoption}

In addition to general applications, we observed that NLP for EBM has been applied within specific medical specialties. Here, we summarized common specialties featured in the papers, such as oncology for conditions like Non-small cell lung cancer (NSCLC) and cardiovascular events such as heart failure. Other diseases are detailed in Supplementary Table \ref{tab:studies}.

\paragraph{Oncology.}\label{oncology}

Cancer is a central topic in EBM, as it demands continuous integration of new research findings to guide evidence-based decisions for accurate diagnosis, effective treatment, and long-term patient management. 
\citet{Saiz2021-vk} introduced Watson Oncology Literature Insights (WOLI), an AI system, by automatically identifying, prioritizing, and extracting relevant oncology research, which facilitated the translation of evidence into clinical practice. Similarly, the Clinical Trial Matching (CTM) system \cite{Alexander2020-bu} was evaluated at a cancer center in Australia with an overall accuracy of 92\% for screening lung cancer patients. These tools highlight how AI-driven systems are increasingly embedded in hospital workflows.




\paragraph{Cardiology.}

Cardiology demands robust evidence to support the decision due to the high prevalence and the critical consequences of diagnostic errors, which can result in severe harm or loss of life. 
For example, the hybrid model proposed by \citet{278} exemplifies clinical practice by automating patient eligibility assessment directly within clinical workflows. In a real-world application on a dataset of 40,000 patients across several clinical care pathways, such as heart failure with reduced and preserved ejection fraction and atrial fibrillation, this model was deployed and achieved an impressive accuracy of 87.3\%. 


\subsection{Clinical trial design and identification}

Not all medical specialties are fully addressed by current research, and even in those with significant focus, the integration of findings into real-world guidance remains insufficient. Automating clinical trial procedures is critical for instant reaction to pandemics or public health emergencies. A crucial step in advancing future clinical trials or experiments is the design phase, where NLP plays a pivotal role. Effective clinical trial design involves structuring and optimizing trials to ensure they align with patient needs and research objectives. NLP tools can enhance the efficiency of clinical trial design by facilitating the automated matching of patients to suitable trials and ensuring trials are aligned with the right patient cohorts. This capability supports a more effective and timely deployment of research resources in emergency health situations.

Eligibility matching and cohort identification is a process of matching patients to clinical trials based on their eligibility and identifying groups of patients (cohorts) who meet specific criteria for inclusion in clinical trials. There are several applications. \citet{Vydiswaran2019-id} proposed a hybrid approach to identify patient cohorts for clinical trials, which combines pattern-based, knowledge-intensive, and feature-weighting techniques to determine if patients meet specific selection criteria. \citet{Segura-Bedmar2019-kj} explored the use of deep learning models for cohort selection, framing it as a multi-label classification task. By employing CNNs and RNNs to process free-text eligibility criteria, this method allows for automatic learning of representations directly from text. Building on these foundations, \citet{Liu2022-il} developed Criteria2Query (C2Q) to extract and transform free-text eligibility criteria into structured, queryable data for cohort identification. More recently, \citet{Murcia2024-sz} proposed the ``TrialMatcher'' algorithm to match veterans for clinical trials using existing information within EHRs. It extracted attributes from patient profiles and eligibility criteria from trial profiles and compared them using the Sørensen-Dice Index (SDI). These applications show the potential of streamlining the process of recruitment and improving future clinical trial design. Now, researchers try to add LLMs to the studies. LLMs like GPT-3.5 or GPT-4 enhance clinical trial workflows by processing complex natural language data, such as patient profiles and trial eligibility criteria. The examples include AutoTrial~\cite{Wang2023-ss}, focusing on trial design, specifically generating eligibility criteria using multi-step reasoning and hybrid prompting and TrialGPT~\cite{Jin2024-cf}, implementing a comprehensive framework for large-scale patient-trial matching, emphasizing real-world deployment and time-saving efficiency.

\subsection{Drug repurposing}

Another frontier application in this field is drug repurposing, which utilizes NLP to analyze existing medical literature and uncover new therapeutic applications for established drugs. By automating the analysis of large datasets such as clinical trials and research papers, NLP speeds up the identification of potential treatments, offering a faster and more cost-effective alternative to traditional drug discovery methods. During the COVID-19 pandemic, there is an urgent need for drugs for treatment. To quickly meet this requirement, the CovidX Network Algorithm~\citet{Gates2020-zn} was developed, which utilized NLP to analyze vast COVID-19 biomedical literature. It ranked potential drug candidates for repurposing, highlighting NLP's power in automating and accelerating evidence synthesis during critical times. Alzheimer’s disease (AD), a progressive neurodegenerative disorder, remains a major global health challenge with limited treatment options and no definitive cure. Despite significant investment in drug development, the failure rate for Alzheimer's-specific drugs in clinical trials remains exceedingly high. To address this, \citet{Daluwatumulle2022-as} employed knowledge graph embeddings to predict AD drug candidates by linking textual data and generating hypotheses from unstructured information. 

\subsection{Question Answering}\label{question-answering}
While EBM is taught according to the five steps: ask, acquire, appraise, apply, and evaluate, a recent trend of application with the advancement in LLMs focuses on treating the entire process as a question-answering (QA) task. \citet{Xie2023-qq} experimented with the consultation of rhinoplasty questions to ChatGPT, which pre-learned knowledge and summarized texts to respond, testing the potential of LLMs to offer valuable feedback. 
Moreover, \citet{Mohammed2024-dx} added bootstrapping to BioBERT and BioGPT so that they could better understand PICO questions from physicians and find potential answers from publications. Expanding on this trend, \citet{Chuan2021-np} introduced Chatbot SOPHIA, which helps users understand their eligibility for clinical trials by answering questions based on trial criteria. Addressing rare cancers, \citet{Jang2022-tj} fine-tuned SAPBERT for QA and NER tasks, ultimately summarizing potential drugs ranked by relevance, such as bevacizumab, temozolomide, lomustine, and nivolumab.

\section{EBM Benchmark dataset}\label{benchmark-dataset}

Here, we summarize the benchmarks used in NLP and EBM (Supplementary Table \ref{tab:dataset}).
%
The tasks frequently involved with these benchmarks are Evidence Retrieval, Evidence Extraction, and Clinical Trial Identification. There is a notable gap in datasets specifically tailored for Evidence Synthesis and Appraisal, as well as Question Answering. The existing datasets are often built upon general texts rather than medical-specific content. For example, CNN-DailyMail~\cite{CNNdaily} is used for Evidence Summarization, but it is not medical-related. We also noticed that the primary data sources for these benchmarks are scholarly articles from PubMed and clinical trials.

\section{Challenges and Future Directions}\label{discussion-future-direction-and-limitations}


EBM is an important, rewarding, and dynamic field that organizes current data to improve healthcare decision-making. By integrating the best available evidence with a healthcare professional's experience and the patient's values, EBM aims to optimize health outcomes. Our focus here is on retrieving, extracting, appraising, synthesizing, and summarizing evidence from biomedical literature such as clinical trials, cohort studies, and case reports. However, conducting these analyses can be both demanding and time-consuming. In this study, we explore key NLP techniques that can streamline and facilitate this process.

Our review indicates that NLP-based systems or pipelines have achieved impressive results in EBM, such as extracting entities like PICO, enhancing the information retrieval engines, automating the evidence synthesis, assessing evidence quality, ranking the evidence with the highest confidence, summarizing the information, and answering questions.
At the same time, as in any other evolving area, there remain challenges ahead. For example, generative models in EBM tasks have demonstrated impressive fluency and scalability, yet their tendency to hallucinate facts, lack source attribution, and sensitivity to prompt phrasing remain significant limitations for clinical use. A core challenge is the validation and trustworthiness of generated outputs, especially in high-stakes domains like medicine. Mechanisms such as RAG offer potential mitigations but require further development and evaluation.

From another perspective, particularly in handling diseases with limited literature or annotated data \cite{Ge2023-da}. Few-shot learning holds significant potential, as it enables models to generalize effectively from a small number of examples, reducing the dependency on large, annotated datasets. This data-efficient approach is crucial for EBM tasks in under-researched areas, such as rare diseases, where annotated resources are scarce. Few-shot learning can help these models adapt quickly to specific clinical needs, allowing for more accurate information extraction, question answering, and evidence synthesis, even with minimal training data.

Additionally, there is a pressing need for more benchmark datasets, especially for Evidence Synthesis and Appraisal and Question Answering. Current resources often rely on general corpora rather than those specifically oriented toward medical content, limiting the development of specialized NLP applications. Researchers can consider and build more meaningful datasets. Moreover, NLP-based tools have not yet been widely applied across all medical specialties, such as Urology and Hepatology, indicating room for expansion in these areas.

Another future direction for NLP in EBM involves incorporating real-world data from various sources, such as mobile devices, social media, and genomics. These data sources capture rich and diverse information beyond traditional clinical records, offering valuable insights into patient behaviors, lifestyle, environmental factors, and genetic predispositions. For example, data from mobile health apps and wearable devices can provide real-time health metrics. At the same time, social media posts may reveal patient self-reported outcomes or experiences that are often missed in clinical settings. Integrating genomic data adds another layer, enabling family history and personalized genomic code into disease risk and treatment response.

Furthermore, the ``black box" nature of many NLP models limits their interpretability and accountability. Biases within training data can restrict NLP's effectiveness and fairness across diverse patient demographics. Additionally, the high computational demands and the need for domain expertise in both NLP and healthcare are resource-intensive. 

To fully realize the potential of NLP for EBM in real-world clinical workflows often involve interdisciplinary scenarios that span multiple conditions, comorbidities, and patient subpopulations. To address these complexities, NLP systems for EBM must evolve toward more holistic, adaptable frameworks capable of reasoning across diverse clinical questions and integrating heterogeneous data sources.

Addressing these limitations is important for enabling efficiency and ultimately contributing to a safer, more equitable healthcare landscape.

\section{Conclusion}\label{conclusion}

Our comprehensive review of over 600 papers resulted in the selection of 129 studies that focus on critical aspects of NLP within EBM. We first provide an overview of EBM, followed by a survey of NLP methods and techniques that address each step of the EBM process. We also explore use cases that demonstrate the application of EBM in various scenarios. Additionally, we review popular datasets and benchmarks. Finally, we present open challenges and future directions for research in this field.
As NLP technologies evolve, they offer promising prospects for harnessing vast amounts of unstructured data, thus supporting clinical and research applications. 

\section*{Limitations}
Our study primarily focuses on English-language publications, potentially overlooking important research published in other languages. The inclusion criteria may have excluded studies indirectly related to EBM and NLP that could provide valuable insights. Additionally, our analysis only covers articles published between 2019 and 2024, which may have led to the omission of significant earlier works that contributed to the foundation of this field. Furthermore, the databases and search engines used in this review are limited, and it is possible that some relevant studies on NLP for EBM during the specified period were not identified.

\section*{Acknowledgments}

This project was sponsored by the National Library of Medicine grants R01LM014344 and R01LM014573.

\bibliography{main_camera_ready}

\appendix

\setcounter{table}{0}
\setcounter{figure}{0}
\renewcommand\figurename{Supplementary Figure} 
\renewcommand\tablename{Supplementary Table}
\section{Appendix}

{
\onecolumn
\subsection{Included Studies}
\footnotesize
\definecolor{c1}{HTML}{E64B35}
\definecolor{c2}{HTML}{4DBBD5}
\definecolor{c3}{HTML}{00A087}
\definecolor{c4}{HTML}{3C5488}
\definecolor{c5}{HTML}{F39B7F}
\definecolor{c6}{HTML}{8491B4}
\definecolor{c7}{HTML}{91d1c2}
\definecolor{c8}{HTML}{dc0000}
\definecolor{c9}{HTML}{7e6148}
\definecolor{c10}{HTML}{b09c85}
\definecolor{c11}{HTML}{F7F7F7}
\definecolor{c12}{HTML}{6EB43F}
\definecolor{c13}{HTML}{F8981D}

\newcommand{\abssum}{\Circled[fill color=c6,inner color=black]{S}\xspace}
\newcommand{\trial}{\Circled[fill color=c10,inner color=black]{T}\xspace}
\newcommand{\ent}{\Circled[fill color=c1,inner color=white]{N}\xspace}
\newcommand{\eval}{\Circled[fill color=c4,inner color=white]{E}\xspace}
\newcommand{\evrank}{\Circled[fill color=c13,inner color=black]{L}\xspace}
\newcommand{\extsum}{\Circled[fill color=c7,inner color=black]{S}\xspace}
\newcommand{\fndsyn}{\Circled[fill color=c11,inner color=black]{Y}\xspace}
\newcommand{\ir}{\Circled[fill color=c5,inner color=black]{I}\xspace}
\newcommand{\quaass}{\Circled[fill color=c12,inner color=black]{A}\xspace}
\newcommand{\qa}{\Circled[fill color=c3,inner color=black]{Q}\xspace}
\newcommand{\rel}{\Circled[fill color=c2,inner color=black]{R}\xspace}

\rowcolors{2}{}{lightgray!30} 
\begin{xltabular}{\columnwidth}{ 
>{\raggedright\arraybackslash}p{3.5cm} 
>{\raggedright\arraybackslash}X 
>{\raggedright\arraybackslash}X
l} 
\caption{Overview of Included Studies. P~- Precision, R~- Recall, Acc~- Accuracy, NDCG~- Normalized Discounted Cumulative Gain.\\
\abssum~- Abstractive Summarization, \trial~- Clinical Trial Design, \ent~- Entity Extraction and Classification, \eval~- Evaluation of Performance, \evrank~- Evidence Ranking and Screening, \extsum~- Extractive Summarization, \fndsyn~- Evidence Synthesis, \ir~- Information Retrieval, \quaass~- Quality Assessment, \qa~- Question Answering, \rel~- Relation Extraction.\label{tab:studies}}\\
\toprule
Study & Model & Disease & Task \\ 
\midrule
\endfirsthead

\multicolumn{4}{l}{{\bfseries \tablename\ \thetable{} -- continued from previous page}} \\
\toprule
Study & Model & Disease & Task  \\ 
\midrule
\endhead

\rowcolor{white}
\midrule 
\rowcolor{white}\multicolumn{4}{r}{Continued on next page} \\ 
\endfoot

\bottomrule
\endlastfoot
\citet{540} & RNN & -- & \trial \ent \eval \\ 
\citet{Alexander2020-bu} & Statistical & Lung Cancer & \trial \ent \eval \\
\citet{Alodadi2019} & Rules & -- & \ent \eval \ir \rel \\
\citet{Aum2021-fp} & Transformer & Cognitive Impairment & \ent \eval \evrank \rel \\

\citet{377} & LLM & -- & \trial \ent \eval \\ 
\citet{Beck2020-bm} & Rules & Breast Cancer & \trial \ent \eval \\
\citet{292} & Rules & Cancer & \ent \rel \\
\citet{Brassey2021-ua} & Rules & -- & \ent \fndsyn \quaass \\
\citet{13} & Transformer & -- & \ent \evrank \\ 
\citet{458} & Random Forest, Logistic LASSO & Rheumatoid Arthritis & \trial \eval \\ 
\citet{CT-EBM-SP} & Transformer & -- & \ent \eval \\
\citet{Chen2019-ze} & Rules & -- & \abssum \trial \ent \eval \rel \\
\citet{487} & CNN & Myocardial infarction & \trial \ent \eval \\ 
\citet{316} & Rules & -- & \trial \ent \eval \\ 
\citet{Chuan2021-np} & CNN & Cancer &  \trial \ent \eval \qa \\
\citet{Cunningham2024-we} & Transformer & Heart Failure & \trial \ent \eval \\

\citet{Daluwatumulle2022-as} & Graph & Alzheimer's Disease (AD) & \ent \eval \ir \\
\citet{380} & LLM & -- & \ent \eval \evrank \\ 
\citet{Devi2024-av} & LLM & Non-Small Cell Lung Cancer (NSCLC) & \trial \ent \eval \\
\citet{DeYoung2021-ql} & Transformer & -- & \abssum \ent \eval \rel \\
\citet{626} & Graph, Statistical & Coronary heart disease, Chest pain, Bronchitis & \eval \qa \\
\citet{Do2024-bl} & Rules, Statistical & Cancer & \trial \ent \\
\citet{Dobbins2022-et} & Transformer & -- & \trial \ent \eval \rel \\
\citet{352} & Transformer, Rules & -- & \trial \ent \eval \ir \rel \\
\citet{399} & Statistical, Learning to Rank & -- & \ent \eval \fndsyn \quaass \\ 
\citet{54} & Transformer & -- & \ent \eval \\
\citet{453} & Rules, Statistical & -- & \ent \eval \\ 
\citet{300} & SVM, CNN, RNN, Transformer & Viral Infection & \trial \ent \eval \\
\citet{Du2021-qx} & Graph & COVID-19 & \eval \ir \\
\citet{369} & Rules & -- & \trial \ent \\ 
\citet{Gates2020-zn} & Graph, Statistical & SARS & \ent \evrank \\
\citet{Ghosh2024-km} & LLM & -- & \ent \eval \\
\citet{Ghosh2024-is} & Transformer, RNN & -- & \ent \eval \\
\citet{Gorska2024-qy} & Transformer, LLM  & -- & \ent \evrank \fndsyn \\
\citet{Gulden2019-dv} & Graph & -- & \ent \eval \extsum \\
\citet{69} & LLM & Peyronie Disease & \ent \eval \ir \quaass \\ 
\citet{Hamed2023-jr} & LLM & Diabetic Ketoacidosis & \abssum \eval \qa \\
\citet{396} & Support Vector Machine (SVM), Random Forest (RF), Logistic Regression (LR), and Stochastic Gradient Descent (SGD), RNN & -- & \trial \ent \eval \ir \\ 
\citet{402} & Graph, CRF, Statistical & Spinal Cord Injury & \trial \ent \eval \fndsyn \\ 
\citet{Hu2023-zc} & Transformer & COVID-19, AD & \ent \eval \\
\citet{15} & LLM & Glaucoma & \eval \qa \\ 
\citet{Jang2022-tj} & Transformer & Recurrent Glioblastoma & \ent \eval \qa \\
\citet{Jiang2024-zi} & LLM & -- & \abssum \eval \qa \\
\citet{5} & Transformer & -- & \eval \ir \quaass \\ 
\citet{Jin2024-cf} & Transformer, LLM & -- & \trial \eval \evrank \ir\\
\citet{Johnston2024-od} & Vector Space Model, Statistical & l-DOPA-induced dyskinesia in Parkinson & \ent \evrank \\
\citet{Kamath2021-ux} & Statistical & -- & \trial \evrank \ir \\
\citet{289} & SVM, Naive Bayes, Random Forest & Epilepsy & \trial \ent \eval \\ 
\citet{Kang2019-rl} & RNN & -- & \ent \eval \\
\citet{2} & Transformer, RNN & -- & \ent \eval \\
\citet{Kang2023-ei}  & Transformer  & COVID-19  & \abssum \ent \eval \fndsyn \\
\citet{345} & SVM & Pediatric Leukemia & \trial \ent \eval \evrank \\ 
\citet{492} & BERT & Cancer & \ent \eval \\ 
\citet{468} & Statistical & Chronic Rhinosinusitis with Nasal Polyps & \ent \\ 
\citet{414} & LR, NB, kNN, SVM, CNN, RNN, FastText, Transformer, ERNIE & Hepatocellular Carcinoma & \trial \ent \eval \\ 
\citet{257} & Transformer & -- & \trial \ent \eval \\ 
\citet{692} & Graph & Cancer & \ent \ir \\ 
\citet{693} & MMR & Cancer & \trial \evrank \ir \\ 
\citet{Kury2020-tc} & Rules & -- & \trial \eval \\
\citet{Lalitha2023-tk} & T5(Text-to-Text Transfer Transformer), BART (Bidirectional Auto-Regressive Transformer) and PEGASUS (Pre-training with Extracted Gap-sentences for Abstractive Summarization Sequence-to-sequence) & -- & \abssum \eval \\
\citet{Lan2024-yd} & Transformer & -- & \ent \eval \quaass \\
\citet{430} & RNN & Cancer & \trial \ent \eval \\ 
\citet{707} & Vector Space model & -- & \evrank \ir \\ 
\citet{393} & Rules & -- & \ent \\ 
\citet{Li2022-kv} & Transformer & -- & \trial \ent \eval \\
\citet{46} & Transformer & Stroke, Colorectal Cancer, Coronary Heart Disease, Heart Failure, Chronic Obstructive Pulmonary Disease, Diabetes, Diabetic Nephropathy, Osteoarthritis, Obesity, Rheumatoid Arthritis, and Diarrhea & \ent \eval \\ 
\citet{451} & RNN & -- & \ent \eval \ir \\ 
\citet{311} & Rules & -- & \trial \ent \eval \rel \\ 
\citet{Liu2022-il} & Rules, RNN & -- &  \trial \ent \rel \\
\citet{Lokker2023-tx} & Transformer & -- & \eval \ir \quaass \\
\citet{154} & Statistical & -- & \ent \\ 
\citet{marshall2023pilot-f} & Transformer & COVID-19 & \ent \fndsyn \ir \\
\citet{268} & GRU, CRF, RNN, Transformer & -- & \ent \eval \rel \\
\citet{375} & Rules & -- & \trial \ent \evrank \\ 
\citet{40} & Statistical & Craniofacial Abnormalities & \ent \eval \\ 
\citet{239} & Graph & -- & \ent \ir \\
\citet{Mohammed2023-mt} & Graph & -- & \ent \eval \extsum\\
\citet{Mohammed2024-dx} & LLM & -- & \ent \eval \fndsyn \qa \\ 
\citet{Murcia2024-sz} & Transformer & -- & \trial \ent \\
\citet{Mutinda2022-jx} & Rules, Statistical, Transformer & Breast cancer & \ent \eval \fndsyn \\
\citet{388} & Transformer & - & \ent \eval \quaass \\ 
\citet{55} & Rules, Statistical & -- & \evrank \ir \\ 
\citet{Newbury2023-hg} & Rules & -- & \ent \eval \\
\citet{750} & Transformer, RNN, SVM & -- & \evrank \ir \\ 
\citet{514} & Rules & Respiratory Tract Infection, Traumatic Brain Injury, and Serious Bacterial Infections & \trial \ent \eval \\ 
\citet{341} & LLM & -- & \trial \eval \\ 
\citet{Norman2019-nm} & Statistical & -- & \evrank \\
\citet{421} & Random Forest, XGBoost & Acquired Pressure Injuries & \ent \eval \\ 
\citet{Pan2021-eg} & Transformer, Graph & COVID-19 & \ent \eval \extsum \qa \rel \\
\citet{Ramprasad2023AutomaticallySummarizing} & Transformer, Longformer & -- & \abssum \ir \\
\citet{MediGPT} & LLM & -- & \eval \qa \\ 
\citet{Rybinski2020-rs} & Rules & -- & \ir \\
\citet{Rybinski2020-mz} & Transformer & -- & \ent \eval \evrank \ir \\
\citet{381} & Knowledge & -- & \ent \ir \\ 
\citet{Saiz2021-vk} & Gradient-boosted Trees & Cancer & \trial \ent \eval \evrank \\
\citet{182} & Statistical & Autism & \eval \ir \\ 
\citet{Sanchez-Graillet2022-up} & Transformer, Schema & Glaucoma, Type 2 diabetes mellitus & \ent \eval \\ 
\citet{Sarker2020-uh} & Statistical & COVID-19 & \eval \extsum \\
\citet{Segura-Bedmar2019-kj} & CNN, RNN & -- & \trial \eval \\
\citet{Shiraishi2024-la} & LLM & Blepharoptosis & \eval \qa \\
\citet{463} & SVM, Logistic Regression, Naive Bayes, Gradient Tree Boosting, Rules, Decision Trees, Random Forests & -- & \trial \ent \eval \\ 
\citet{324} & Transformer, RNN & -- & \ent \eval \\ 
\citet{Stylianou2021-ek} & Transformer & -- & \ent \eval \\
\citet{Tian2021-oh} & Transformer & -- & \trial \ent \quaass \\ 
\citet{332} & Statistical, Transformer & AD & \trial \ent \eval \\ 
\citet{389} & Rules & Organ dysfunction in septic shock & \trial \ent \eval \ir \\ 
\citet{Tsubota2022-qf} & Transformer & -- & \ent \eval \ir \\
\citet{278} & Rules& Cardiovascular Events & \trial \eval \\ 
\citet{310} & Rules & -- & \ent \eval \\ 
\citet{Unlu2024-oj} & LLM & Heart Failure & \eval \qa \\
\citet{274} & Rules & mRNA Cancer & \ent \ir \\ 
\citet{Vydiswaran2019-id} & Rules & -- & \trial \ent \eval \\
\citet{Wang2022-xs} & Transformer & -- & \ent \eval \ir \\
\citet{269} & Statistical, Transformer & -- & \ent \eval \\ 
\citet{Wang2023-ss} & LLM & -- & \trial \eval \fndsyn \rel \\
\citet{Wang2024-rx} & CNN, Graph, Rules & Hepatocellular Carcinoma & \ent \eval \\
\citet{Witte2024-bk} & Transformer & Glaucoma, Type II Diabetes & \ent \eval \\
\citet{Xie2022-oa} & Transformer & -- & \ent \extsum \\
\citet{Xie2023-qq} & LLM & Rhinoplasty & \qa \\
\citet{493} & LLM & -- & \abssum \eval \fndsyn \quaass \\ 
\citet{Xu2023-au} & Graph & Cancer & \ent \evrank \quaass \\
\citet{Yang2023-cn} & Transformer & Cancer & \trial \ent \\
\citet{438} & Not Specify & Atrial Fibrillation & \trial \ent \eval \\ 
\citet{240} & Transformer & -- & \eval \quaass \\ 
\citet{371} & LLM & -- & \trial \ent \eval \\ 



\citet{425} & Transformer & -- & \ent \quaass \\ 
\citet{860} & Transformer, RNN & -- & \ent \eval \\ 
\citet{zhang2024span-based-h} & Transformer & COVID-19, AD & \ent \eval \ir \\
\citet{Zheng2024-wa} & LLM & Glaucoma & \eval \qa \\

\end{xltabular}
\twocolumn}


{
\onecolumn
\subsection{Benchmark Dataset}
\footnotesize
\rowcolors{2}{}{lightgray!30} 
\begin{xltabular}{\textwidth}{
>{\raggedright\arraybackslash}p{20ex}l
>{\raggedright\arraybackslash}X
l
>{\raggedright\arraybackslash}X
}
\caption{Overview of recent benchmark datasets. P - Population. I - Intervention. C - Comparison. O - Outcome~~\cite{picodef}. RCT - Randomized controlled trial. ETC - Anything that doesn’t fit into the
categories above. GPG - GNU Privacy Guard. CMS - Content Management System.}
\label{tab:dataset}\\
\toprule
Dataset & Avail. & Label & Annotation & Description\\
\midrule
\endfirsthead

\multicolumn{5}{l}%
{{\bfseries \tablename\ \thetable{} -- continued from previous page}} \\
\toprule
Dataset & Avail. & Label & Annotation & Description\\
\midrule
\endhead

\rowcolor{white}
\midrule 
\rowcolor{white}\multicolumn{5}{r}{Continued on next page} \\
\endfoot

\bottomrule
\endlastfoot
Alzheimer’s disease RCT~\cite{Hu2023-zc} & Public & P, I, C, O & Manual & 150 Alzheimer’s disease RCT abstracts \\
Chia~\cite{Kury2020-tc} & Public & Non-query-able, Post-eligibility, Informed consent, Pregnancy considerations, Parsing error, Non-representable, Competing trial, Context error, Subjective judgment, Not a criteria, Undefined semantics, Intoxication considerations & Manual & Eligibility statements from 1000 clinical trials and the dataset includes 12,409 annotated eligibility criteria \\
Clinical trials on cancer~\cite{MediGPT} & Private & Eligible, Not eligible & Manual & 6M eligibility statements in clinical trials \\
COVID-19 corpus~\cite{Hu2023-zc} & Public & P, I, C, O & Manual & 150 COVID-19 RCT abstracts \\
CT\nobreakdash-EBM\nobreakdash-SP~\cite{CT-EBM-SP} & Public & Anatomy, pharmacological and chemical substances, pathologies, and lab tests, diagnostic or  therapeutic procedures & Manual & 1,200 texts about clinical trials with entities \\
EBM\nobreakdash-COMET~\cite{Ghosh2024-is} & Public & Physiological or clinical, Death, Life impact, Resource use, Adverse events & Manual & 300 RCT abstracts \\
EBM\nobreakdash-NLP~\cite{Nye2018-zs} & Public & P, I, O & Manual & 4,993 medical abstracts from literatures on PubMed \\
EliIE~\cite{EliIE} & Public & Condition, observation, drug/substance, and procedure or device & Manual & 230 Alzheimer’s disease RCT documents \\
GGPONC~\cite{GGPONC} & Public & Recommendation creation date, Type of recommendation, Recommendation grade, Strength of consensus, Total vote in percentage, Literature references, Expert opinion, Level of evidence, Edit State & Automatic & 25 GPGs with 8,414 text segments from the CMS\\
LCT~\cite{Dobbins2022-et} & Public & Clinical, Demographic, Logical, Qualifiers, Temporal and Comparative, Other & Manual & 1,000+ eligibility statements \\
Limsi\nobreakdash-Cochrane dataset~\cite{Limsi-Cochrane} & Public & Systematic reviews, Included studies, Data   forms, Text entries, Excluded studies, Diagnostic tests, Test results, Study IDs, Numerical, Summary scores & Manual & 1,939 meta-analyses from 63 systematic reviews of diagnostic test accuracy from the Cochrane Library \\
MedReview~\cite{Zhang2024} & Private & medical related topics (e.g. Wounds, Urology) & Manual & Meta-analysis results and narrative summaries from the Cochrane Library \\
MS$^{\wedge}$2~\cite{DeYoung2021-ql} & Public & Background, Goal, Methods,   Detailed findings, Further study, Recommendation, Evidence quality, effect, ETC & Manual & 470k documents and 20K summaries from the scientific literature \\
NICTA\nobreakdash-PIBOSO~\cite{NICTA-PIBOSO} & Public & P, I, O, Background, Study Design, Other & Manual & 1,000 biomedical abstracts \\
PICO\nobreakdash-Corpus~\cite{PICO-corpus} & Public & P, I, C, O & Manual & 1,011 breast cancer RCT abstracts\\
RedHOT~\cite{Ghosh2024-is} & Public & P, I, O & Manual & 22,000 social media posts from Reddit spanning 24 health conditions \\
Illness dataset~\cite{MediGPT} & Private & Alzheimer’s, Parkinson’s, Cancer, and Diabetes domains & Manual & 22,660 tweets \\
Symptom2Disease dataset~\cite{MediGPT} & Private & 24 diseases, each described by 50 symptom profiles & Manual & 1,200 data points \\
Trialstreamer~\cite{marshall2020trialstreamer-z} & Public & P, I, O, RCT classifiers & Manual & 191 RCT publications \\
\end{xltabular}
\twocolumn
}

\newpage
\newpage
\onecolumn
\subsection{Queries}
\label{sec:query}

\begin{verbatim}
{nlp_keywords} = natural language processing
    OR nlp
    OR language model
    OR large language model
    OR llm
    OR computational linguistics
    OR information extraction
    OR information retrieval
    OR clinical trial retrieval
    OR text summarization
    OR question answering
    OR sentence segmentation
    OR ner
    OR named entity recognition
    OR tokenization)
{ebm_keywords) = evidence-based medicine
    OR ebm
    OR evidence-based practice
    OR ebp
    OR clinical trial
\end{verbatim}

\paragraph{PubMed}

\begin{verbatim}
(({nlp_keywords}[Title/Abstract]) AND ({ebm_keywords}[Title/Abstract]))
\end{verbatim}


\paragraph{IEEE Xplore}

\begin{verbatim}
(("Abstract":{nlp_keywords}) AND ("Abstract":{ebm_keywords}))
    OR (("Title":{nlp_keywords}) AND (("Title": {ebm_keywords})))
\end{verbatim}

\paragraph{ACM}

\begin{verbatim}
(Abstract:{nlp_keywords} AND Abstract:{ebm_keywords})
    OR (Title:{nlp_keywords} AND Title:{ebm_keywords})
\end{verbatim}


\paragraph{ACL}

\begin{verbatim}
(Abstract:{nlp_keywords} AND Abstract:{ebm_keywords})
    OR (Title:{nlp_keywords} AND Title:{ebm_keywords})
\end{verbatim}



\end{document}